\documentclass[letterpaper]{article}

\usepackage{aaai19, times, helvet, courier}
\usepackage{graphicx, subfigure, svg, tikz, pgfplots}
\usepackage{amssymb, amsmath, bbm}
\pgfplotsset{compat=1.14}
\usetikzlibrary{arrows.meta, matrix, decorations.pathreplacing}

\usepackage[utf8]{inputenc} 
\usepackage[T1]{fontenc}    
\usepackage{url}            
\usepackage{booktabs}       
\usepackage{amsfonts}       
\usepackage{nicefrac}       
\usepackage{microtype}      
\usepackage{todonotes}      
\usepackage{amsmath, amssymb, graphicx, braket}
\usepackage{caption}

\DeclareMathOperator*{\argmax}{arg\,max}

\title{Debate Dynamics for Human-comprehensible Fact-checking on Knowledge Graphs\thanks{This is a short workshop paper. The conference paper can be downloaded from https://arxiv.org/abs/2001.00461.
}}

\author{%
  Marcel Hildebrandt\thanks{These authors contributed equally to this work.}, Jorge Andres Quintero Serna\textsuperscript{$\dagger$}, Yunpu Ma, \AND  Martin Ringsquandl, Mitchell Joblin, Volker Tresp \\
  \\
  Siemens Corporate Technology, Munich, Germany\\
  \texttt{\{firstname.lastname\}@siemens.com} \\
}

\if false
\author{%
  David S.~Hippocampus\thanks{Use footnote for providing further information
    about author (webpage, alternative address)---\emph{not} for acknowledging
    funding agencies.} \\
  Department of Computer Science\\
  Cranberry-Lemon University\\
  Pittsburgh, PA 15213 \\
  \texttt{hippo@cs.cranberry-lemon.edu} \\
}

\fi 
\begin{document}
\maketitle
\begin{abstract}
We propose a novel method for fact-checking on knowledge graphs based on debate dynamics.  The underlying idea is to frame the task of triple classification as a debate game between two reinforcement learning agents which extract arguments -- paths in the knowledge graph -- with the goal to justify the fact being true (thesis) or the fact being false (antithesis), respectively. Based on these arguments, a binary classifier, referred to as  the judge, decides whether the fact is true or false. The two agents can be considered as sparse feature extractors that present interpretable evidence for either the thesis or the antithesis. In contrast to black-box methods, the arguments enable the user to gain an understanding for the decision of the judge. Moreover, our method allows for interactive reasoning on knowledge graphs where the users can raise additional arguments or evaluate the debate taking common sense reasoning and external information into account. Such interactive systems can increase the acceptance of various AI applications based on knowledge graphs and can further lead to higher efficiency, robustness, and fairness. 

\end{abstract}
\section{Introduction}
\label{sec:introduction}

\if false
\begin{itemize}
    \item WHAT IS A KG?
    \item FOR WHICH TASKS ARE KGs EMPLOYED (GOOGLE KG, Q\&A, RECOMMENDATIONS, DB)
    \item LINK PREDICTION, SCORING BASED ON EMBEDDINGS
    \item OPEN AI 
    \item BOUNDARY TO THE IBM STUFF ON DEBATES
    \item STRESS IMPORTANCE OF TRANSPARENCY, INTERACTION WITH USERS (DATA CURATION), BIASES
\end{itemize}
\fi


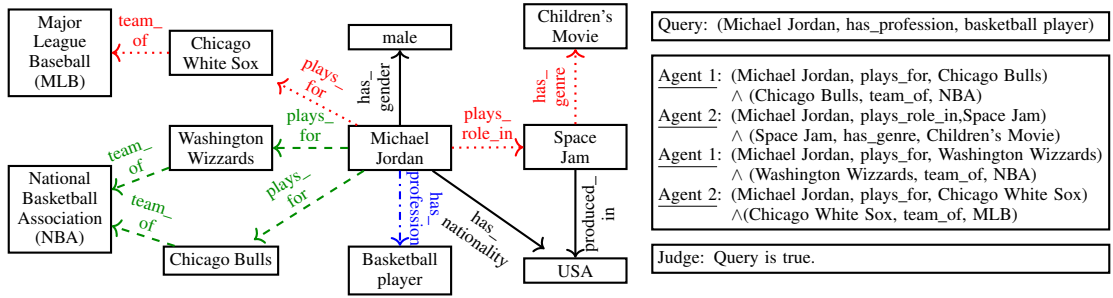
\begin{figure*}
\begin{center}
        \begin{tikzpicture}[thick,scale=1.2,every node/.style={scale=0.68}]
    	\tikzset{state/.style={ draw, minimum width=2cm}}
    	\node[state] (MJ)[align=center] {Michael \\ Jordan};
    	\node[state] (WW) [left=of MJ, align=center]{Washington \\Wizzards};
    	\node[state] (CB) [below=of WW]{Chicago Bulls};
    	\node[state] (NBA) [left=of WW, align=center, yshift=-1.2cm, xshift=0.35cm]{National \\ Basketball \\ Association \\ (NBA)};
    	\node[state] (CWS) [above=of WW, yshift=-0.5cm, align=center]{Chicago \\White Sox};
    	\node[state] (MLB) [left=of CWS, align=center, xshift=0.35cm]{Major \\League \\ Baseball \\ (MLB)};
    	\node[state] (male) [above=of MJ]{male};
        \node[state] (BP) [below=of MJ, align=center]{Basketball \\ player};
    	\node[state] (USA) [right=of BP, xshift=-0.07cm]{USA};
    	\node[state] (SJ) [above=of USA, align=center, yshift=0.25cm]{Space \\Jam};
    	\node[state] (CM) [above=of SJ, align=center]{Children's\\ Movie};

\node[draw,text width=8.7cm] (Q) [right=of CM, xshift=-1cm] {Query: (Michael Jordan, has\_profession, basketball player) };
\node[draw,text width=8.7cm, ] (D) [below=of Q, yshift=1.2cm] {

\underline{Agent 1}: (Michael Jordan, plays\_for, Chicago Bulls)  \\ $\quad \quad \quad \; \; \,$ $\wedge$  (Chicago Bulls, team\_of, NBA) \\

\underline{Agent 2}: (Michael Jordan, plays\_role\_in,Space Jam)   \\ $\quad \quad \quad \; \; \,$ $\wedge$ (Space Jam, has\_genre, Children's Movie)

\underline{Agent 1}: (Michael Jordan, plays\_for, Washington Wizzards)\\ $\quad \quad \quad \; \; \,$ $\wedge$ (Washington Wizzards, team\_of, NBA)

\underline{Agent 2}: (Michael Jordan, plays\_for, Chicago White Sox) \\ $\quad \quad \quad \; \; \,$ $\wedge$(Chicago White Sox, team\_of, MLB)
};

\node[draw,text width=8.7cm] (J) [below=of D, yshift=1.2cm] {Judge: Query is true. };

    	\path[->, shorten >=0.15cm, dashed, black!45!green, align=center] (MJ) edge node[sloped, above] {plays\_\\for} (CB);
    	\path[->, dashed, black!45!green, align=center] (MJ) edge node[sloped, above] {plays\_\\for} (WW);
    	\path[->, dashed, black!45!green, align=center] (CB) edge node[sloped, above] {team\_\\of} (NBA);
    	\path[->, dashed, black!45!green, align=center] (WW) edge node[sloped, above] {team\_\\of} (NBA);
    	\path[->, shorten >=0.2cm, dotted, red, align=center] (MJ) edge node[sloped, above] {plays\_\\for} (CWS);
    	\path[->, dotted, red] (CWS) edge node[sloped, above, align=center] {team\_\\of} (MLB);
    	\path[->, align=center] (MJ) edge node[sloped, above] {has\_\\gender} (male);
    	\path[->, dotted, red, align=center] (MJ) edge node[sloped, above] {plays\_\\role\_in} (SJ);
    	\path[->, shorten >=0.2cm, align=center] (MJ) edge node[sloped, below] {has\_\\nationality} (USA);
    	\path[->, dotted, red, align=center] (SJ) edge node[sloped, above] {has\_\\genre} (CM);
    	\path[->, align=center] (SJ) edge node[sloped, below] {produced\_\\in} (USA);

    	\path[->, align=center, dash dot, blue] (MJ) edge node[sloped, above] {has\_\\profession} (BP);
    	\end{tikzpicture}
\end{center}

	\hspace{1em}
	\label{fig:kg_tensor}
\caption{The agents debate whether Michael Jordan is a professional basketball player. While agent 1 extracts arguments from the KG promoting that the fact is true (green), agent 2 argues that it is false (red). Based on the arguments the judge decides that  Michael Jordan is a professional basketball player.}
\label{fig:MJ_debate}
\end{figure*}

Knowledge graphs (KG) are multi-relational, graph-structured databases which store facts about the real world. Thereby, entities correspond to vertices and binary relations to edge types that specify how the entities are related to each other. KGs are useful resources for various artificial intelligence (AI) tasks in different fields such as named entity disambiguation in NLP \cite{han2010structural} or visual relation detection in computer vision \cite{baier2017improving}. Examples of large sized KGs include Freebase  \cite{bollacker2008freebase}, YAGO \cite{suchanek2007yago}, and WordNet \cite{miller1995wordnet}. In particular, the Google Knowledge Graph \cite{KGblogpost} is a well-known example of a commercial KG with more than 18 billion facts, used by the search engine to enhance results. One major issue, however, is that most real-world KGs are incomplete (i.e., true facts are missing) or contain false facts. Machine learning algorithms designed to solve this problem try to infer missing triples or detect false facts based on observed connectivity patterns. 

Most machine learning approaches for reasoning on KGs embed both entities and predicates into low dimensional vector spaces. Then a score for the plausibility of a triple can be computed based on these embeddings. Common to most of these methods is their black-box nature because what contributed to this score is hidden from the user. This lack of transparency constitutes a potential limitation when it comes to deploying KGs in real world settings. Explainability as a development focus for new machine learning models has gained attention in the last few years (see \cite{ribeiro2016should}, \cite{mahendran2015understanding}, or \cite{montavon2017explaining}) as the success of machine learning and AI continues, while the ability to understand and interpret the increasingly complex and powerful methods staggers behind. Laws that require explainable algorithms and models have been considered and implemented \cite{goodman2017european}. 
 Additionally, in contrast to one-way black-box configurations, comprehensible machine learning methods allow to build systems where both machines and users interact and influence each other. 

In this work we describe a novel method for triple classification based on reinforcement learning. Inspired by the concept outlined in \cite{irving2018ai} to increase AI safety via debates, we model the task of triple classification as a debate between two agents each presenting arguments for the thesis (the triple is true) and the antithesis (the triple is false), respectively. Based on these arguments, a binary classifier, referred to as the judge, decides whether the fact is true or false. In that sense the two agents act as feature extractors that present evidence for and against the validity of the fact. In contrast to most methods based on representation learning, the arguments can be displayed to the user such that he can audit the classification of the judge and potentially overrule the decision. 
Further, our  method can be used to create tools for knowledge graph curation or question answering where the users can interact with the system by inputting additional arguments or make a decision based on the agents' arguments. Moreover, mining evidence for both the thesis and the antithesis can be considered as adversarial feature generation, making the classifier more robust towards contradictory evidence or corrupted data. To the best of our knowledge, our method constitutes the first  model based on debate dynamics for triple classification in KGs.


\section{Background and Related Work}
\label{sec:related_work}
\if false
\begin{itemize}
 \item KGs
 \item  EMBEDDING BASED METHODS (TRANSLATIONAL; FACTORIZATION; CONVE)
 \item  PATHS-BASED METHODS (PATH RANK; DEEP WALK; MINERVA)
 \item definition of fact prediction and link prediction
 \end{itemize}
\fi 
 
In this section we provide a brief introduction to KGs in a formal setting and review the most relevant related work. Let $\mathcal{E}$ denote the set of entities and consider the set of binary relations $\mathcal{R}$. A knowledge graph $\mathcal{KG} \subset \mathcal{E} \times \mathcal{R} \times \mathcal{E} $ is a collection of facts stored as triples of the form $(s, p, o)$ – subject, predicate, and object. To indicate whether a  triple is true or false, we consider the binary characteristic function $ \phi : \mathcal{E} \times \mathcal{R} \times \mathcal{E} \rightarrow \{ 1, 0 \}$. For all $(s,p,o) \in \mathcal{KG}$ we assume $\phi(s,p,o) = 1$ (i.e., a KG is a collection of true facts). However, in case a triple is not contained in $\mathcal{KG}$, it does not imply that the corresponding fact is false but rather unknown (open world assumption). Triple classification (or fact-checking) is concerned with  predicting the truth value $ \phi ( s,p,o )$.

Representation learning is an effective and popular technique underlying many KG reasoning methods. The basic idea is to project both entities and relations into a low dimensional vector space. Then the likelihood of triples is modelled in terms of a functional on the embedding spaces. Popular methods based on representation learning include the translational embedding method TransE \cite{bordes2013translating}, factorization approaches such ComplEx \cite{trouillon2016complex}, or deep neural network methods including ConvE \cite{dettmers2018convolutional}. While these methods efficiently encode the local neighborhood of entities, they cannot reason through more complex inference paths in KGs. Therefore, path-based reasoning methods have been studied. For instance, the Path-Ranking Algorithm (PRA) proposed in \cite{lao2011random} uses a combination of weighted random walks through the graph for inference. \cite{wenhan_emnlp2017} proposed the reinforcement learning based path searching approach called DeepPath, where an agents picks relational paths between entity pairs. Recently, and more related to our work, \cite{minerva} proposed the multi-hop reasoning method MINERVA. The basic idea is to display the query subject and predicate to the agent and let the agent perform a policy guided walk to the correct object entity. The paths that MINERVA produces also lead to some degree of explainability. However, we find that only actively mining arguments for the thesis and the antithesis, thus exposing the user to both sides of a debate, allows to make a well-informed decision. 

Our method is inspired on the concept outlined in \cite{irving2018ai} where the authors propose the use of debate dynamics for the AI alignment task. Their concept is based on training agents to play a zero sum debate game where they raise arguments which are evaluated by a judge. More concretely, given a question or a proposed action, two or multiple agents take turns making short statements until a human can decide which of the agents gave the more convincing information. While a major part of their work consists of theoretical considerations, they also conduct an initial experiment on MNIST with an automated judge: First, an image is shown to two agents and each of them selects a digit. Then, the two agents sequentially select pixels of the image with the goal to convince a sparse classifier that the pixels stem from an image that displays the digits to which they are committed. Thereby, the classifier cannot observe the complete image but only the pixels revealed by the agents. The authors state that their eventual goal is to produce natural language debates, where humans judge dialogues between the agents. In that regard, our method can be seen as a first step in that direction since paths in a knowledge graph correspond to factual statements that can be easily translated to natural language.

\section{Our Method}
\label{sec:our_method}

\begin{figure*}
 \begin{center}
    \includegraphics[width=1.\textwidth]{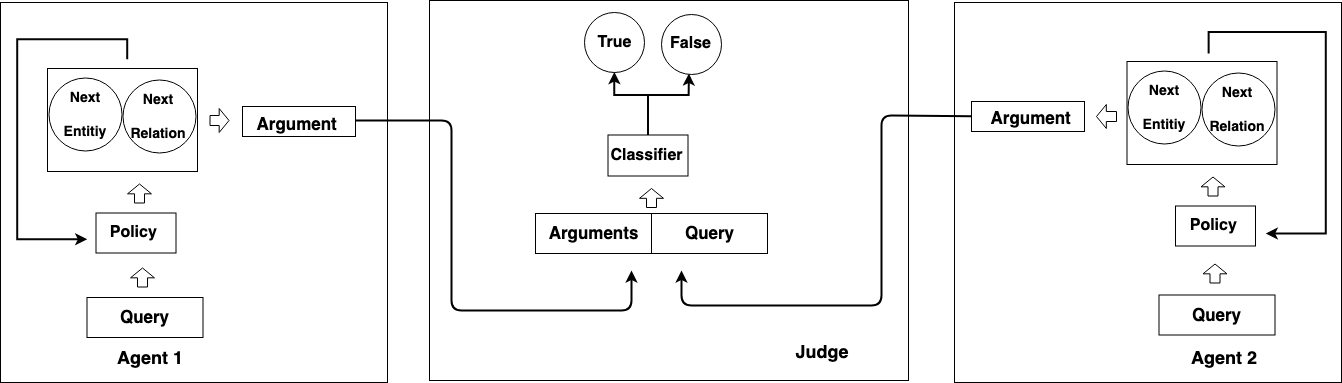}
  \caption{Sketch of the architecture of our method: The two agents extract arguments from the KG. Along with the query relation and the query object, these arguments are processed by the judge who classifies whether the query is true or false. 
  }
  \label{fig:architecture}   
 \end{center}
\end{figure*}

We formulate the task of fact prediction in terms of a debate between two opposing agents. Thereby, a query triple corresponds to the statement around which the debate is centered. The agents mine paths on the KG that serve as evidence for the thesis or the antithesis. More concretely, they traverse the graph sequentially and select the next hop based on a policy that takes previous transitions and the query triple into account. The transitions are added to the current path extending the argument. 
All paths are processed by a binary classifier called the judge that tries to distinguish between true and false triples.  An exemplary debate is shown in Figure \ref{fig:MJ_debate}. The main steps of a debate can be summarized as follows:
\begin{enumerate}
    \item A query triple around which the debate is centered is presented to both agents
    \item The two agents take turns extracting paths from the KG that serve as arguments for the thesis and the antithesis.
    \item The judge processes the arguments along with the query triple and estimates the truth value of the query triple.
\end{enumerate}
While the parameters of the judge are fitted in a supervised fashion, both agents are trained to navigate through  the  graph  using reinforcement learning. More concretely, their environments are modelled via the fixed horizon, deterministic Markov decision processes outlined below. 

\paragraph{States}
The fully observable state space $\mathcal{S}$ for each agent is given by $\mathcal{E}^2 \times \mathcal{R} \times \mathcal{E}$. Intuitively, we want the state to encode the query triple $q = (s_q,p_q,o_q)$  and the location of exploration $e^{(i)}_t$ (i.e., the current location) of agent $i \in \{1,2 \}$ at time $t$. Thus, a state $S^{(i)}_t \in \mathcal{S}$ for time $t \in \mathbb{N}$ is represented by $S^{(i)}_t = \left(e^{(i)}_t, q\right)$.

\paragraph{Actions} \sloppy
 The set of possible actions for agent $i$ from a state $S^{(i)}_t = \left(e^{(i)}_t, q\right)$ is denoted by $\mathcal{A}_{S^{(i)}_t}$. It consists of all outgoing edges from the entity $e^{(i)}_t$ and the corresponding target entities. More formally, $\mathcal{A}_{S^{(i)}_t} = \left\{(r,e) \in \mathcal{R} \times \mathcal{E} :  S^{(i)}_t = \left(e^{(i)}_t, q\right) \land \left(e^{(i)}_t,r,e\right) \in \mathcal{KG}\right\}\, .$ Moreover, we denote with $A_t^{(i)} \in \mathcal{A}_{S^{(i)}_t}$ the action that agent $i$ performed at time $t$. We include self-loops for each node such that the agent can stay at the current node.

\paragraph{Environments}
The environments evolve deterministically by updating the state according to the agents' actions (i.e., by changing the agents' locations). The query fact remains the same. Formally, the transition function of agent $i$ at time $t$ is given by $\delta_t^{(i)}({S^{(i)}_t},A_t^{(i)}) := \left(e^{(i)}_{t+1}, q\right)$ with $S^{(i)}_t = \left(e^{(i)}_{t}, q\right)$ and $A_t^{(i)} = \left(r, e^{(i)}_{t+1}\right)$.

\paragraph{Policies}
We denote the history of agent $i$ up to time $t$ with the tuple $H^{(i)}_t = \left(H^{(i)}_{t-1}, A^{(i)}_{t-1}\right)$ for $t \geq 1$ and $H^{(i)}_0 = (q)$. The agents encode their histories via recurrent neural networks. The output is mapped to a distribution over all admissible actions (i.e., transitions). This induces agent specific policies $\pi^{(i)}_\theta: H^{(i)}_t \mapsto \left(p_1, p_2, \dots, p_m\right) \in \mathbb{R}^{m}_{\geq 0}$, where $m = \lvert \mathcal{A}_{S^{(i)}_t} \rvert$ and $\theta$ denotes the set of all trainable parameters. Then, the next action of agent $i$ is sampled according to $
    A_t^{(i)} \sim \text{Categorical}\left(p_1, p_2, \dots, p_m\right)
$.

\begin{table*}
\center
\resizebox{\textwidth}{!}{
\begin{tabular}{ c c | c}
\hline
 \textbf{Query:}
  &  Richard Feynman $\xrightarrow{nationality}$ USA? & Nelson Mandela $\xrightarrow{hasProfession}$ Actor? \\  
  \hline
  
\textbf{Agent 1:} 
 & Richard Feynman $\xrightarrow{livedInLocation}$ Queens &  Nelson Mandela $\xrightarrow{hasFriend}$ Naomi Campbell \\
 & $\wedge$ Queens $\xrightarrow{locatedIn}$ USA & Naomi Campbell  $\xrightarrow{hasDated}$ Leonardo DiCaprio \\
   \hline
\textbf{Agent 2:} & Richard Feynman $\xrightarrow{hasEthnicity}$ Russian people  & Nelson Mandela $\xrightarrow{hasProfession}$ Lawyer \\
 & $\wedge$ Russian people$\xrightarrow{geographicDistribution}$ Republic of Tajikistan  & $\wedge$ Lawyer $\xrightarrow{\_specializationOf}$ Barrister 
  \vspace{0.3cm}

\end{tabular}
}
\caption{Two examples of debates generated by our method: While agent 1 argues that the query triple is true and agent 2 argues that it is false. The underscore indicates inverse relations.}
\label{tab:deb_example}
\end{table*}

\paragraph{Debate Dynamics}
In a first step, the query triple $q$ with truth value $\phi(q) \in \{0,1\}$  is presented to both agents. Agent 1 argues that the fact is true, while agent 2 argues that it is false. Similar to most formal debates, we consider a fixed number of rounds $N \in \mathbb{N}$. In every round $n = 1,2, \dots, N$ the agents start graph traversals with fixed length $T \in \mathbb{N}$ from the subject entity of the query $s_q$. More concretely, agent 1 starts generating a sequence $\tau_n^{(1)}$ of length $T-1$ consisting of entities and relations followed by a mandatory end of argument action. Then agent 2 proceeds by producing a similar sequence $\tau_n^{(2)}$ starting from $s_q$ ending with the stop action. After $N$ rounds the debate terminates and concatenating all paths leads to the overall sequence of arguments $\tau$. The judge processes $\tau$ and predicts the truth value of the triple. This induces a binary classifier $f(\tau) \in (0,1)$. The parameters of the judge are trained in a supervised fashion to approximate the underlying truth value of the triple $\phi (q)$. An overview of the architecture of our method is depicted in Figure \ref{fig:architecture}.


\if false
Algorithm  \ref{alg:debate} contains a pseudocode of our method at inference time.
\begin{algorithm}
\SetAlgoLined
\DontPrintSemicolon
\SetKwInOut{Query}{input}
\SetKwInOut{Judge}{input}
\SetKwInOut{AgentY}{input}
\SetKwInOut{AgentN}{input}
\SetKwInOut{Output}{output}
\Query{Triple query $q=(s_q,p_q,o_q)$}
\Output{Classification score $t \in [0,1]$ of the judge along with the list of arguments $L$}
 $L\leftarrow [ \; ]$$\;$\tcp{Initialize the list of all arguments}
 \tcp{Loop over the debate rounds}
 \For{$n = 1$ to N}{ 
 \tcp{Loop over the two agents}
    \For{$i = 1$ to 2}{
     $e_1^{(i)} \leftarrow s_q$$\;$\tcp{Initialize the position of the agent}
     $A \leftarrow [ \; ]$$\;$\tcp{Initialize the current argument}
     \tcp{Loop over the path}
     \For{$t = 1$ to T }{
        Sample a transition $(r,e)\sim \pi_\theta^{(i)}$ from $e_t^{(i)}$ \;
        $A\text{.append}(r,e)\;$ \tcp{Extend the argument}
        $e_{t+1}^{(i)} \leftarrow e\;$ \tcp{Update the position of the agent}
     }
     $L\text{.append}(A)\;$ \tcp{Extend the list of all argument}}
 }
 Process $L$ by the judge and compute the score $t$ \;
 \KwRet{$L$ and $t$}
 \caption{Debate dynamics at inference time}
 \label{alg:debate}
\end{algorithm}
\fi

Since agent 1 argues that the thesis is true, the rewards $R^{(1)}_n$ after every round $n \leq N$ are given by the classification score $R^{(1)}_n = f\left(\tau_n^{(1)}\right)$. Similarly, the rewards of agent 2 are given by $R^{(2)}_n = - f\left(\tau_n^{(2)}\right)$. Intuitively, this means that the agents receive high rewards whenever they extract an argument that is considered by judge as strong evidence for their position. To find the best set of parameters the agents maximize the expected cumulative rewards where standard methods such as  REINFORCE \cite{williams1992simple} can be employed.

\if false
\begin{equation}
    R^{(i)}_n = \begin{cases}
 t &\text{if } i = 1 \\
-t &\text{otherwise.}
\end{cases}
\end{equation}

An overview of the overall architecture of R2D2 is depicted in Figure \ref{fig:architecture}. A pseudocode of R2D2 at inference time is shown in Algorithm 1 in the supplementary material.
\begin{figure*}
 \begin{center}
    \includegraphics[width=\textwidth]{Full_Architecture.png}
  \caption{The overall architecture of R2D2: The two agents extract arguments from the KG. Along with the query relation and the query object, these arguments are sequentially processed by the judge who classifies whether the query is true or false. 
  }
  \label{fig:architecture}   
 \end{center}
\end{figure*}

\paragraph{Reward Maximization}
The agents maximize the expected cumulative reward given by
\begin{equation}
    G^{(i)} := \sum_{n = 1}^N  R^{(i)}_n \, .
\end{equation}
Thus, the agents' maximization problems are given by
\begin{equation}
\label{eq:objective_agent}
    \argmax_{\theta^{(i)}} \mathbb{E}_{q \sim \mathcal{KG}_+}\mathbb{E}_{\tau^{(i)}_1, \tau^{(i)}_2, \dots, \tau^{(i)}_N \sim \pi_\theta^{(i)}}\left[G^{(i)} \left\vert\vphantom{\frac{1}{1}}\right. q = (s_q, p_q, o_q)\right] \, ,
\end{equation}
where $\mathcal{KG}_+$ is the set of training triples that contain in addition to observed triples in $\mathcal{KG}$ also unobserved triples. The rationale is as follows: As knowledge graphs only contain true facts, simple sampling from a query $(s_q,p_q,o_q) \sim \mathcal{KG}$ would create a dataset without negative labels. This is the reason why it is common to create corrupted triples that are constructed from correct triples $(s,p,o)$ by replacing the object with an entity $\tilde o$ to create a false triple $(s,p,\tilde o) \notin \mathcal{KG}$ (see \cite{bordes2013translating}).

During training the first expectation in Equation \eqref{eq:objective_agent} is substituted with the empirical average over the training set. The second expectations is approximated by the empirical average over multiple rollouts to reduce the sample variance. To maximize Equation \eqref{eq:objective_agent} various methods such as REINFORCE \cite{williams1992simple} can be employed.

SIMPLIFIED ARCHITECTURE FIGURE

\fi

\section{Discussion}
\label{sec:discussion}

In this section we present the results of a preliminary experiment and discuss applications and directions of future works. We implemented our method as outlined in Section \ref{sec:our_method} where the policies of the agents consists of LSTM cells with a dense neural network stacked on top. The judge processes each argument individually by a feed forward neural network, sums the output for each argument up and processes the resulting sum by a linear, binary classifier. Both agents and the judge have separate lookup tables that contain embeddings for the relations and entities. We run this first experiment on the benchmark dataset FB15k-237 \cite{toutanova2015representing} which contains around 15,000 different entities mainly focused around culture, sports, and politics. Following \cite{socher2013reasoning} we select the following relations for training and testing purposes: 'gender', 'profession', 'nationality', 'ethnicity', and 'religion'. Thereby we found that 82.4\% of the query triples in the balanced test set (i.e., equally many true and false facts) were classified correctly by the judge. While these results are promising for a first experiment, we leave the majority of the empirical work including experimenting with different architectures to future research.

By manually examining the quality of the extracted paths we found mostly reasonable arguments (e.g. when arguing that a person has a certain nationality, the agents frequently bring up that the fact the person is born in a city in this country). However, there were also some arguments that do not make intuitive sense. We conjecture that this is partially due to the fact that the embeddings of entities encode their neighborhoods. While the judge has access to this information through the training process, it remains hidden to the user. For example, when arguing for the fact that Nelson Mandela was an actor (see Table \ref{tab:deb_example}) 
the argument of agent 1 requires the user to know that both Naomi Campbell as well as Leonardo DiCaprio are actors (which is encoded in FB15k-237). Then this argument serves as evidence that Nelson Mandela was also an actor since people tend to have friends that share their profession (social homophily). However, without this context information it is not intuitively clear why this is a reasonable argument. 
To asses whether the arguments are informative for users in an objective setting we plan to conduct a survey where respondents take the role of the judge making a classification decision based on the agents' arguments. 

Many KGs, including DBpedia, Freebase, and Wikidata, are (among other techniques) semiautomatically built and maintained with the assistance of manual efforts to ensure that the knowledge keeps expanding and is kept up-to-date and accurate \cite{ge2016visualizing}. While manually validating triples is laborious, most machine learning methods for this task are based on a numerical confidence score which is hard to interpret. Our method can be employed for interactive KG curation and fact checking. More concretely, if there are doubts concerning the truthfulness of a triple, the user can examine the arguments to make an informed decision or input arguments that the agents might have missed. Suppose, for example,  the triple $(Barack Obama, nationality, USA)$ is classified as false and the user observes that one agent brought up the argument that Barack Obama was born in Kenya -- a fact most likely mined from an unreliable source from the web. The user can then enter the correct fact that Obama is born in Hawaii which results in the correct classification. 
On a related note, \cite{nickel2015review} raise the point that when applying statistical methods to incomplete KGs  the results are likely to be affected by biases in the data generating process and should be interpreted accordingly. Otherwise, blindly following  recommendations from KG reasoning methods can even strengthen these biases. While the judge in our method also exploits skews in the underlying data, the arguments of the agents can help to identify these biases and potentially exclude problematic arguments from the classification decision. Moreover, our method can be integrated into QA systems. Instead of simply providing an answer to a question, the system could then also display evidence for and against this answer allowing the user to trace back the decision. This added transparency can potentially increasing user acceptance. 


\section{Conclusion}
\label{sec:conclusion}

We have described a novel approach for triple classification in KGs based on a debate game between two opposing reinforcement learning agents that argue whether a fact is true or false. The two agents search the KG for arguments that shift the judge, a binary classifier, towards their position. Since the judge bases its decision on the extracted arguments, the user can examine the arguments and trace back the classification decision. This also allows for building tools where the user interacts with the system leading to higher acceptance of KG reasoning tools. While the first experiments that we conduct in the scope of this work are promising, significant further research, including experiments with different architectures for the agents and the judge, is required. Another directions for future work is a systematic, qualitative analysis of the arguments.

\newpage
\bibliographystyle{aaai}
\bibliography{bibliography}

\begin{thebibliography}{}

\bibitem[\protect\citeauthoryear{Baier, Ma, and
  Tresp}{2017}]{baier2017improving}
Baier, S.; Ma, Y.; and Tresp, V.
\newblock 2017.
\newblock Improving visual relationship detection using semantic modeling of
  scene descriptions.
\newblock In {\em International Semantic Web Conference},  53--68.
\newblock Springer.

\bibitem[\protect\citeauthoryear{Bollacker \bgroup et al\mbox.\egroup
  }{2008}]{bollacker2008freebase}
Bollacker, K.; Evans, C.; Paritosh, P.; Sturge, T.; and Taylor, J.
\newblock 2008.
\newblock Freebase: a collaboratively created graph database for structuring
  human knowledge.
\newblock In {\em Proceedings of the 2008 ACM SIGMOD international conference
  on Management of data},  1247--1250.
\newblock AcM.

\bibitem[\protect\citeauthoryear{Bordes \bgroup et al\mbox.\egroup
  }{2013}]{bordes2013translating}
Bordes, A.; Usunier, N.; Garcia-Duran, A.; Weston, J.; and Yakhnenko, O.
\newblock 2013.
\newblock Translating embeddings for modeling multi-relational data.
\newblock In {\em Advances in neural information processing systems},
  2787--2795.

\bibitem[\protect\citeauthoryear{Das \bgroup et al\mbox.\egroup
  }{2018}]{minerva}
Das, R.; Dhuliawala, S.; Zaheer, M.; Vilnis, L.; Durugkar, I.; Krishnamurthy,
  A.; Smola, A.; and McCallum, A.
\newblock 2018.
\newblock Go for a walk and arrive at the answer: Reasoning over paths in
  knowledge bases using reinforcement learning.
\newblock In {\em ICLR}.

\bibitem[\protect\citeauthoryear{Dettmers \bgroup et al\mbox.\egroup
  }{2018}]{dettmers2018convolutional}
Dettmers, T.; Minervini, P.; Stenetorp, P.; and Riedel, S.
\newblock 2018.
\newblock Convolutional 2d knowledge graph embeddings.
\newblock In {\em Thirty-Second AAAI Conference on Artificial Intelligence}.

\bibitem[\protect\citeauthoryear{Ge \bgroup et al\mbox.\egroup
  }{2016}]{ge2016visualizing}
Ge, T.; Wang, Y.; de~Melo, G.; Li, H.; and Chen, B.
\newblock 2016.
\newblock Visualizing and curating knowledge graphs over time and space.
\newblock {\em Proceedings of ACL-2016 System Demonstrations}  25--30.

\bibitem[\protect\citeauthoryear{Goodman and
  Flaxman}{2017}]{goodman2017european}
Goodman, B., and Flaxman, S.
\newblock 2017.
\newblock European union regulations on algorithmic decision-making and a
  “right to explanation”.
\newblock {\em AI Magazine} 38(3):50--57.

\bibitem[\protect\citeauthoryear{Han and Zhao}{2010}]{han2010structural}
Han, X., and Zhao, J.
\newblock 2010.
\newblock Structural semantic relatedness: a knowledge-based method to named
  entity disambiguation.
\newblock In {\em Proceedings of the 48th Annual Meeting of the Association for
  Computational Linguistics},  50--59.
\newblock Association for Computational Linguistics.

\bibitem[\protect\citeauthoryear{Irving, Christiano, and
  Amodei}{2018}]{irving2018ai}
Irving, G.; Christiano, P.; and Amodei, D.
\newblock 2018.
\newblock Ai safety via debate.
\newblock {\em arXiv preprint arXiv:1805.00899}.

\bibitem[\protect\citeauthoryear{Lao, Mitchell, and
  Cohen}{2011}]{lao2011random}
Lao, N.; Mitchell, T.; and Cohen, W.~W.
\newblock 2011.
\newblock Random walk inference and learning in a large scale knowledge base.
\newblock In {\em Proceedings of the Conference on Empirical Methods in Natural
  Language Processing},  529--539.
\newblock Association for Computational Linguistics.

\bibitem[\protect\citeauthoryear{Mahendran and
  Vedaldi}{2015}]{mahendran2015understanding}
Mahendran, A., and Vedaldi, A.
\newblock 2015.
\newblock Understanding deep image representations by inverting them.
\newblock In {\em Proceedings of the IEEE conference on computer vision and
  pattern recognition},  5188--5196.

\bibitem[\protect\citeauthoryear{Miller}{1995}]{miller1995wordnet}
Miller, G.~A.
\newblock 1995.
\newblock Wordnet: a lexical database for english.
\newblock {\em Communications of the ACM} 38(11):39--41.

\bibitem[\protect\citeauthoryear{Montavon \bgroup et al\mbox.\egroup
  }{2017}]{montavon2017explaining}
Montavon, G.; Lapuschkin, S.; Binder, A.; Samek, W.; and M{\"u}ller, K.-R.
\newblock 2017.
\newblock Explaining nonlinear classification decisions with deep taylor
  decomposition.
\newblock {\em Pattern Recognition} 65:211--222.

\bibitem[\protect\citeauthoryear{Nickel \bgroup et al\mbox.\egroup
  }{2015}]{nickel2015review}
Nickel, M.; Murphy, K.; Tresp, V.; and Gabrilovich, E.
\newblock 2015.
\newblock A review of relational machine learning for knowledge graphs.
\newblock {\em Proceedings of the IEEE} 104(1):11--33.

\bibitem[\protect\citeauthoryear{Ribeiro, Singh, and
  Guestrin}{2016}]{ribeiro2016should}
Ribeiro, M.~T.; Singh, S.; and Guestrin, C.
\newblock 2016.
\newblock Why should i trust you?: Explaining the predictions of any
  classifier.
\newblock In {\em Proceedings of the 22nd ACM SIGKDD international conference
  on knowledge discovery and data mining},  1135--1144.
\newblock ACM.

\bibitem[\protect\citeauthoryear{Singhal}{2012}]{KGblogpost}
Singhal, A.
\newblock 2012.
\newblock Introducing the knowledge graph: things, not strings.
\newblock [Online; accessed 28-March-2019].

\bibitem[\protect\citeauthoryear{Socher \bgroup et al\mbox.\egroup
  }{2013}]{socher2013reasoning}
Socher, R.; Chen, D.; Manning, C.~D.; and Ng, A.
\newblock 2013.
\newblock Reasoning with neural tensor networks for knowledge base completion.
\newblock In {\em Advances in neural information processing systems},
  926--934.

\bibitem[\protect\citeauthoryear{Suchanek, Kasneci, and
  Weikum}{2007}]{suchanek2007yago}
Suchanek, F.~M.; Kasneci, G.; and Weikum, G.
\newblock 2007.
\newblock Yago: a core of semantic knowledge.
\newblock In {\em Proceedings of the 16th international conference on World
  Wide Web},  697--706.
\newblock ACM.

\bibitem[\protect\citeauthoryear{Toutanova \bgroup et al\mbox.\egroup
  }{2015}]{toutanova2015representing}
Toutanova, K.; Chen, D.; Pantel, P.; Poon, H.; Choudhury, P.; and Gamon, M.
\newblock 2015.
\newblock Representing text for joint embedding of text and knowledge bases.
\newblock In {\em Proceedings of the 2015 Conference on Empirical Methods in
  Natural Language Processing},  1499--1509.

\bibitem[\protect\citeauthoryear{Trouillon \bgroup et al\mbox.\egroup
  }{2016}]{trouillon2016complex}
Trouillon, T.; Welbl, J.; Riedel, S.; Gaussier, E.; and Bouchard, G.
\newblock 2016.
\newblock {Complex embeddings for simple link prediction}.
\newblock In {\em International Conference on Machine Learning (ICML)},
  volume~48,  2071--2080.

\bibitem[\protect\citeauthoryear{Williams}{1992}]{williams1992simple}
Williams, R.~J.
\newblock 1992.
\newblock Simple statistical gradient-following algorithms for connectionist
  reinforcement learning.
\newblock {\em Machine learning} 8(3-4):229--256.

\bibitem[\protect\citeauthoryear{Xiong, Hoang, and
  Wang}{2017}]{wenhan_emnlp2017}
Xiong, W.; Hoang, T.; and Wang, W.~Y.
\newblock 2017.
\newblock Deeppath: A reinforcement learning method for knowledge graph
  reasoning.
\newblock In {\em Proceedings of the 2017 Conference on Empirical Methods in
  Natural Language Processing (EMNLP 2017)}.
\newblock Copenhagen, Denmark: ACL.

\end{thebibliography}
\end{document}